\documentclass{article}

\usepackage[preprint,nonatbib]{neurips_2026}

\usepackage[utf8]{inputenc} 
\usepackage[T1]{fontenc}    
\usepackage{hyperref}       
\usepackage{url}            
\usepackage{booktabs}       

\usepackage{amsmath,amssymb}

\usepackage{amsfonts}       
\usepackage{nicefrac}       
\usepackage{microtype}      
\usepackage{xcolor}         

\usepackage{graphicx}
\usepackage{tcolorbox}

\title{How Much Data is Enough? \\
The Zeta Law of Discoverability in Biomedical Data, featuring the enigmatic Riemann zeta function
}

\author{
Paul Thompson\thanks{Webpage: https://enigma.ini.usc.edu} \\
Stevens Institute for Neuroimaging \& Informatics \\
University of Southern California \\
Los Angeles, CA, USA \\
\texttt{pthomp@usc.edu}
}

\begin{document}

\maketitle

\begin{abstract}
  
   \textbf{How much data is enough to make a scientific discovery?} As biomedical datasets grow to millions of individuals and AI models expand to billions of parameters, progress increasingly depends on our ability to predict when additional data will meaningfully improve detection of real effects. In practice, advances in machine learning often rely on trial-and-error benchmarking across models, modalities, and sample sizes, with limited theoretical guidance on when performance should improve substantially and when it will plateau. The same questions keep coming up: \textbf{how much data do we need to make reliable discoveries? How quickly will discoveries emerge as we add more data? How much data do we need to train a diagnostic model with clinically useful accuracy? Do we need better models or better input variables? And how do we know?} Existing scaling laws and sample complexity theory do not adequately explain the highly variable gains observed across domains, particularly in scientific and biomedical settings where high-dimensional signals must be aligned across heterogeneous data types. \\ \\
We propose a unifying framework for cross-modal discoverability based on the spectral structure of data, signals, and their alignment. Many common performance metrics, including AUC, can be expressed in terms of an effective signal-to-noise parameter that accumulates across spectral modes of an encoder and a cross-modal operator, analogous to those used in canonical correlation analysis and vision-language models. Under mild assumptions, the growth of this parameter with sample size follows a zeta-like law governed by the decay rates of the signal spectrum and the covariance spectrum of the data, leading naturally to the appearance of the Riemann zeta function. \\ \\
This perspective yields several insights. Discoverability depends not only on effect size and sample size, but also on the spectral alignment between modalities, which can be learned and optimised through encoder design. Encoder choice re-maps the spectrum, explaining why sparse models, low-rank embeddings, contrastive vision-language models, and other representation learning approaches can boost sample efficiency. Heterogeneity and subtyping can transform diffuse high-rank effects into lower-rank structure, improving scaling behaviour. Unexpectedly, adding a modality such as text or language embeddings may increase discoverability even if the additional information is partially redundant, by steepening the shared spectral decay. \\ \\
The framework also predicts \textbf{cross-over behaviour}: simpler models may perform best at small sample sizes, while higher-capacity or multimodal encoders start to outperform them once sufficient data stabilises the relevant spectral modes. This progression is analogous to the Tower of Hanoi puzzle, where deeper levels of structure become accessible only after earlier structure has stabilised. Likewise, higher-order spectral modes become usable only once sufficient data has stabilised the dominant modes. Such cross-overs help to explain why empirical benchmarking often yields conflicting conclusions about which models perform best. \\ \\
After reviewing classical results including the Davis–Kahan theorem, we illustrate these principles in applications such as multimodal diagnosis of disease, imaging genetics, functional MRI of the brain, and topological data analysis. Together, these results suggest a general law: the success of data scaling depends on the spectral geometry of signals, encoders, feature representations, and cross-modal operators. The resulting \textbf{zeta law of discoverability helps us to predict when adding data, new modalities, better models, or better input features will help the most}. The framework ranks common models - including sparse models such as elastic net, latent variable models such as VAEs, and contrastive pretraining - according to their spectral slopes, clarifying when each should improve data efficiency. As a corollary, the theory provides a form of power analysis for certain classes of high-dimensional machine learning models and suggests principled ways to design representations that accelerate discovery.

\end{abstract}

\section{How Much Data is Enough? }

A common question in biomedical data analysis is whether we have enough data to answer a question. \\ \\  \textbf{As more data comes in, will our predictive models get a lot better with more data, or do we need to design a more ingenious model? Or, put another way: can our questions be answered with the data we have today, and, if not, how much will it help to collect more? } \\ 

Even global research consortia face these questions. Some questions cannot be answered today even by pooling all the available worldwide data, and by using all the best current models. Clearly the type of statistical model matters. Predictive and diagnostic models can be as simple as cut-offs applied to a single medical measure or diagnostic test, to decide whether a person may have a specific condition or disease. Linear classifiers can also combine multiple biomedical measures, and deep networks may read in thousands or millions of inputs, using encoders and transformers to identify reliable connections within and across datasets. Complex approaches often require more data to train, but it is not always clear how much data is enough. Today, many AI and machine learning practitioners use trial and error to identify the best methods, without realising which methods should overtake others as more data comes in. Here we offer a new theory to guide biomedical scientists and AI developers in deciding which methods will overtake others in performance as more data comes in, and how fast performance should improve as they add more data. In parallel, we identify when it helps the most to design better models (Section 5), or design better input features (Section 7). The laws that govern discoverability have a surprising connection to the \textbf{Riemann zeta function}, which appears in perhaps the greatest unsolved problem in mathematics: the Riemann hypothesis. This is why we call them the \textbf{zeta law}.\\

\begin{figure}[ht]
\centering
\includegraphics[width=0.8\linewidth]{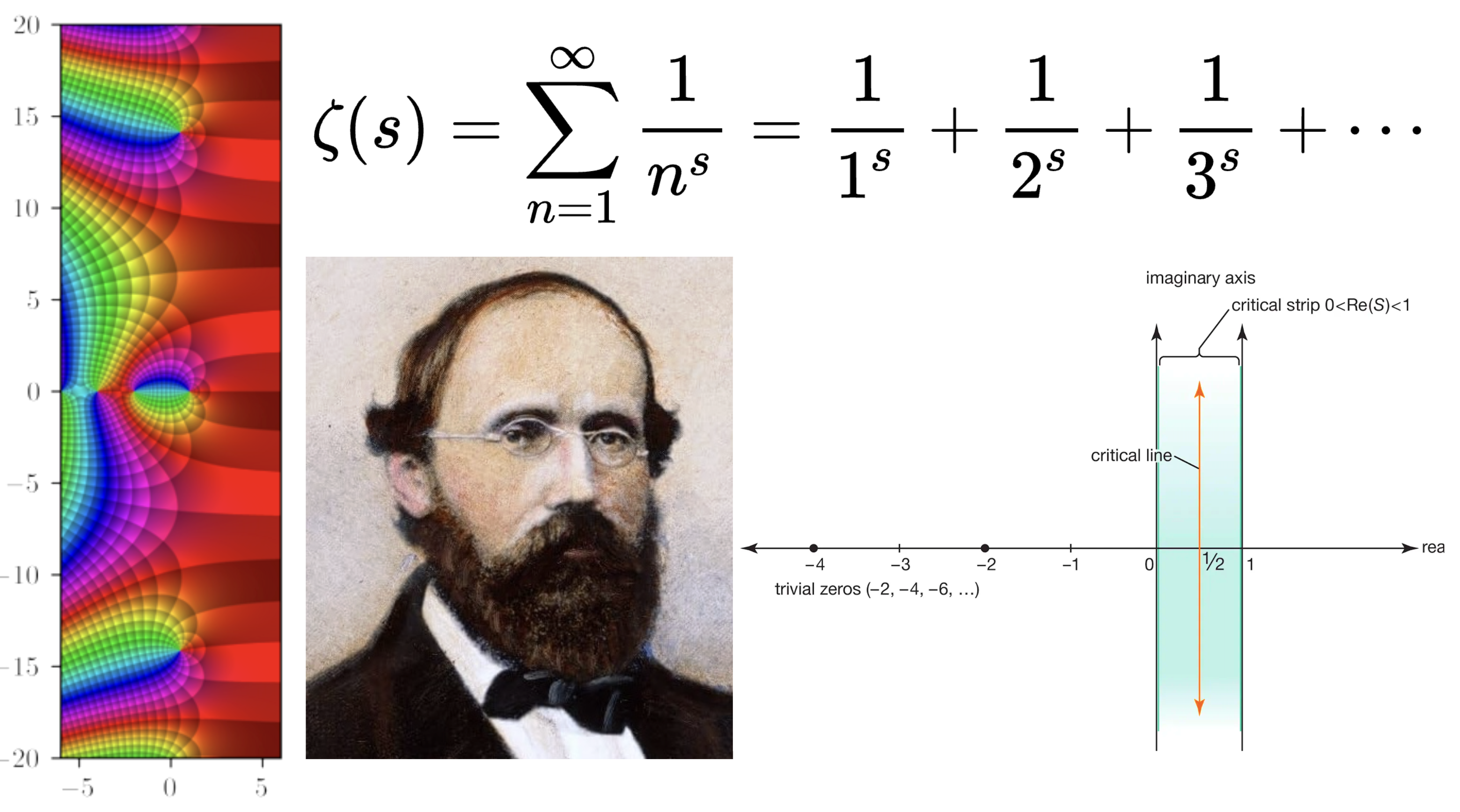}
\caption{
\textbf{Riemann Zeta Function and the Riemann Hypothesis.} The enigmatic Riemann zeta function, whose formula is shown at the top, was introduced by \textbf{Bernhard Riemann (1826--1866)}. It appears in the famous unsolved \textbf{Riemann hypothesis}. The hypothesis states that all nontrivial zeros of $\zeta(s)$ in the complex plane lie on the critical line with real part equal to $1/2$ (drawing courtesy of Encyclopedia Britannica). The colored two-dimensional panel visualizes the complex-valued structure of $\zeta(s)$ in the complex plane, where the color reflects the magnitude and phase of the function and highlights the highly structured pattern of its zeros. As we shall see in this paper, the zeta function also describes how quantities that decay as power laws accumulate across scales. In neuroscience and biomedical data, many signals exhibit a similar power-law structure, where a few dominant modes explain substantial variation but many weaker modes contribute small but important effects. The zeta law builds on this analogy, describing how predictive signal accumulates across identifiable modes as sample size increases, leading us to methods to answer pressing questions more efficiently, and with less data.
}
\label{fig:riemann}
\end{figure}

At an intuitive level, the zeta law describes how discovery improves as the sample size grows, and as progressively weaker patterns in the data become detectable. Many biomedical datasets contain signals that are spread across many “modes” of variation - some strong, others subtle. Here, a \textbf{mode} refers to a pattern consisting of \textbf{many features that tend to vary together}, rather than a single feature - for example, a coherent spatial pattern across brain regions, or a group of genes whose expression levels tend to rise and fall together.  With limited data, only the strongest modes can be reliably estimated. As more samples are collected, noise averages out, and weaker modes can be recovered with confidence. The number of “identifiable modes” (defined later) gradually increases with sample size, at a rate governed by how quickly signal strength decays across the eigenfunction spectrum (the decay rate $\beta$ of disease-aligned spectral energy; see Section 4). \\ 

In many natural systems - natural or medical images, genetic measures, or even brain activity - signal strength often decays approximately as a power law (see Section 4.2). A power law simply means that a small number of patterns explain a lot of the variation, while others contribute successively smaller effects, gradually tapering off rather than stopping suddenly. When this happens, we can estimate how much information we are likely to recover as the sample size grows. If the many smaller effects together add up in a predictable way, we can estimate how much data we need to detect them. In mathematical terms, the cumulative detectable spectral energy follows a form related to the Riemann zeta function. This helps to explain why some machine learning methods perform very well on some prediction problems, such as detecting Alzheimer’s disease from brain images, and how their accuracy improves steadily with more data, while others, such as identifying genetic variants influencing the brain, require dramatically larger sample sizes or improved representations, or cannot be solved at all. \\

This same zeta law suggests some guiding principles for designing encoders and transformers in multimodal learning and predictive modeling (see Section 5.6). Effective representations concentrate biologically meaningful variation into a few dominant modes, increasing useful spectral energy and reducing the sample size needed for discovery. If we are designing a diagnostic classifier, for example, the effective signal corresponds to the cumulative energy of the disease contrast projected onto the identifiable eigenmodes of the covariance operator of the input data at the current sample size. Put simply, it reflects how much of the disease-related variation can be distinguished from normal variation, once we account for noise and limited sample size. This provides a mathematical link between sample size, spectral decay, and achievable classification accuracy. In the biomedical examples below, we often use classification of disease as an example, but the theory applies to other predictive models as well. \\

Cross-modal learning introduces an additional mechanism to improve efficiency. Encoders that align informative “modes” across different data modalities increase the “shared spectral energy” (see Section 5.4), allowing us to find meaningful associations more readily even when the underlying data is unchanged. In practical terms, alignment means learning representations where related patterns from different data types correspond to each other - for example, imaging patterns and gene expression patterns that tend to occur in the same patients are brought into correspondence in a shared representation. This helps explain why multimodal approaches can outperform unimodal ones, and why auxiliary modalities that are not very useful on their own, can greatly improve predictive performance when used in conjunction with others.\\ 

Several surprising consequences follow from this formulation. Adding informative features can reduce sample complexity - the amount of data needed to reach a given predictive accuracy. But adding modalities that are not very useful on their own can greatly improve discoverability (see Section 6.3). Often, an additional modality of data steepens the cross-modal spectrum and increases the “shared spectral energy”. The cross-modal spectrum describes how strongly patterns in different data types correspond to each other, such as imaging patterns that tend to occur with particular genetic or clinical profiles. This alignment depends on the encoder, meaning the mathematical or AI model used to represent each data type in a comparable form. Some encoders emphasize patterns that are more consistent across modalities, concentrating shared information into fewer stronger modes so that meaningful associations can be detected with limited sample sizes. A \textbf{steeper spectrum} means that a few shared patterns stand out more clearly, so we can detect them more easily with limited data. For example, adding a text encoder to a biomedical prediction problem may appear redundant, yet the cross-modal operator may develop larger singular values, concentrating the signal into fewer dominant modes that we can estimate reliably from finite samples. A related effect is seen in deep learning. When we train models to perform a task (such as diagnosing a disease based on input biomedical data), models with many parameters sometimes need \emph{fewer} samples to train than simpler models if they learn a compact representation of the data. In this case, the model organizes the information so that much of the disease-related variation is  compressed into a relatively small number of identifiable patterns, making learning more efficient. When this happens, \textbf{the effective complexity of the problem is lower than it first appears}, improving sample efficiency. Throughout the paper, we return to one central idea: predictive accuracy depends on how much disease-relevant signal is concentrated in the eigenmodes that we can reliably estimate reliably at a given sample size, a general principle that we formulate as the \textbf{zeta law}.\\

We begin with some simple examples to build our intuition for these effects.

\section{Uniform convergence and sample complexity in 1D: Growth Charts }

To illustrate some of these ideas, consider a ``growth chart'' constructed from a reference dataset of healthy individuals, modeling how a biological measure changes across the human lifespan. In children, growth charts are commonly used to determine whether a child’s height or weight falls within the statistically normal range for their age. Suppose we wish to use such a chart to determine whether a new person’s brain measure is unusual relative to the healthy population - for example, whether it lies in the lowest 5\% of the reference distribution, defined by a curve on the chart separating normality from abnormality.

How many individuals are needed in the reference sample to estimate this 5th percentile cut-off with a given degree of accuracy? If we had 1,000 individuals versus 10 million in the reference dataset, how different would the estimated centile be?

A well-developed theory addresses this question. The Dvoretzky--Kiefer--Wolfowitz (DKW) inequality provides a bound on the worst-case difference between the empirical distribution function estimated from data and the true population distribution \cite{dkw1956,massart1990}.  This inequality, proven in 1956 by Aryeh Dvoretzky, Jack Kiefer, and Jacob Wolfowitz, and later sharpened in 1990 by Pascal Massart, is:

\begin{equation}
P\!\left(
\sup_{x \in \mathbb{R}}
\left|
F_n(x) - F(x)
\right|
>
\varepsilon
\right)
\le
2 e^{-2 n \varepsilon^2}
\end{equation}

Here $F_n(x)$ is the empirical cumulative distribution function estimated from $n$ samples, and $F(x)$ is the true population distribution.

This result is an example of a \emph{concentration inequality}: it quantifies how quickly empirical estimates converge to the true distribution as the number of samples increases.

By rearranging this inequality, we can estimate how much data is required to ensure that centiles have stabilized and can be trusted with a specified degree of probability. Many measures of calibration approximately follow a square-root law: estimation error typically decreases in proportion to

\[
\frac{1}{\sqrt{N}},
\]

where $N$ is the sample size used to train the model.

Calibration here means that predicted probabilities or centiles agree with their observed frequencies in the population - for example, in the limit, approximately 5\% of individuals should fall below the estimated 5th percentile. The accuracy of this agreement typically improves at the square-root rate. Doubling the precision therefore requires approximately four times as much data.

This one-dimensional setting illustrates the first intuition behind the zeta law. With finite samples, only a limited portion of the underlying structure can be estimated reliably. As sample size grows, progressively finer structure becomes identifiable, and predictive accuracy improves in a predictable manner.

\section{Convergence of Diagnostic Accuracy (AUC) with Sample Size for a Linear Classifier}

Suppose we want to classify whether a person has a specific disease (for example, Alzheimer's disease) based on a set of features measured from brain MRI, and suppose we can also estimate how these features vary in a large-scale healthy population. For now, consider a simple binary problem: the new person either has the disease or is a healthy control. We measure $p$ different brain features, such as cortical thickness values, regional brain volumes, or connectivity measures. Many of these features are correlated, and each may carry a small amount of information about the disease.

If we examine features one at a time and select the best single predictor, we may obtain only modest diagnostic accuracy. A single brain measure might achieve an AUC of about 0.65 for classifying bipolar disorder or major depression, around 0.8 for anorexia nervosa, and perhaps 0.9 or higher for Alzheimer's disease. These values reflect how separable the disease and control groups are when using only one dimension of information. \emph{But what happens when we combine features into a multivariate classifier, and what happens as we increase the sample size used to train it? How accurate can the classifier ultimately become?}

A simple linear method for combining features is linear discriminant analysis (LDA). In LDA, the optimal direction for separating two groups is the vector connecting the group means, adjusted for the covariance structure of the data. We seek a direction in feature space along which the disease and control groups differ the most, while accounting for correlations among features.

Let the mean feature vector in controls be $\boldsymbol{\mu}_0$, and in the disease group be $\boldsymbol{\mu}_1$. Define the difference vector

\begin{equation}
\mathbf{d} = \boldsymbol{\mu}_1 - \boldsymbol{\mu}_0 .
\end{equation}

If the covariance matrix of the features is $\Sigma$, then the optimal linear discriminant direction is proportional to

\begin{equation}
\mathbf{w} \propto \Sigma^{-1} \mathbf{d}.
\end{equation}

This direction rescales the feature space so that differences along reliable (low-noise) directions contribute more than differences along noisy ones.

The degree of separation between the two groups is given by the Mahalanobis distance,

\begin{equation}
\Delta^2 = \mathbf{d}^T \Sigma^{-1} \mathbf{d}.
\end{equation}

This quantity measures how far apart the class means are when distances are expressed relative to the covariance structure of the data. It can also be interpreted as a measure of abnormality, as the same quadratic form quantifies how unusual an individual’s brain pattern is relative to the control population.

If the data are approximately Gaussian, the AUC of the optimal linear classifier depends directly on this distance:

\begin{equation}
AUC = \Phi\!\left( \frac{\Delta}{\sqrt{2}} \right),
\end{equation}

where $\Phi$ is the cumulative distribution function of the standard normal distribution.

This relationship shows that classification accuracy depends on how much disease signal accumulates across features relative to the covariance of the noise.

\section{How Adding Features Improves AUC}

\subsection{Spectral decomposition of Mahalanobis distance}

To understand how AUC improves as we include more features, it is helpful to express the Mahalanobis distance in the eigenbasis of the covariance matrix. Let

\begin{equation}
\Sigma = U \Lambda U^T
\end{equation}

where the columns of $U$ are eigenvectors and $\Lambda$ contains eigenvalues
$\lambda_1 \ge \lambda_2 \ge \cdots \ge \lambda_p$.

We can decompose the disease contrast vector into this basis:

\begin{equation}
\mathbf{d} = \sum_{k=1}^{p} \alpha_k \mathbf{u}_k .
\end{equation}

The Mahalanobis distance becomes

\begin{equation}
\Delta^2 =
\sum_{k=1}^{p}
\frac{\alpha_k^2}{\lambda_k}.
\end{equation}

Each eigenmode contributes signal proportional to its squared projection onto the disease contrast, divided by the noise variance along that direction.

When training data are limited, we cannot estimate all these eigenmodes reliably. Estimation error in the covariance matrix grows rapidly in higher dimensions, so only the first $K(N)$ modes can be trusted at sample size $N$. As more data are collected, additional modes become stable and contribute to classification accuracy.

We can therefore write an approximate expression for the effective Mahalanobis distance:

\begin{equation}
\Delta^2(N)
\approx
\sum_{k=1}^{K(N)}
\frac{\alpha_k^2}{\lambda_k}.
\end{equation}

As $N$ increases, the number of identifiable modes $K(N)$ grows, allowing more disease-relevant variation to contribute to prediction accuracy.

This explains why AUC improves with sample size even when the underlying biology is unchanged. More modes of variation can be estimated reliably from the available data.

\subsection{Connection to spectral decay and the zeta law}

In many biological contexts, both signal strength and noise variance follow approximate power laws across modes:

\begin{equation}
\frac{\alpha_k^2}{\lambda_k}
\sim
k^{-\beta}.
\end{equation}

Power law structure has been observed in eigenspectra of neural activity, resting-state fMRI connectivity, and effect sizes in genome-wide association studies, where a small number of dominant components explain substantial variation but many smaller effects persist across a long tail.

Under this assumption, cumulative Mahalanobis distance grows as

\begin{equation}
\Delta^2(N)
\sim
\sum_{k=1}^{K(N)}
k^{-\beta}.
\end{equation}

This partial sum has the form of a \textbf{truncated Riemann zeta function}. As $K(N)$ grows with sample size, diagnostic accuracy improves according to the rate at which spectral energy accumulates.

Different diseases may have different spectral profiles:

Disorders with strong low-order modes can achieve high AUC with modest sample sizes. Neurological conditions such as Alzheimer's disease, epilepsy, and stroke follow this pattern. Disorders whose signal is distributed across many weak modes require larger datasets before classification accuracy improves substantially. Psychiatric disorders such as schizophrenia, bipolar disorder, and major depression may fall into this category.

Representations that concentrate disease signals into earlier modes effectively steepen spectral decay, increasing accuracy at finite sample size.

This perspective helps explain empirical observations across many neuroimaging studies. Alzheimer's disease often shows high AUC with moderate sample sizes because disease signal is concentrated in relatively few dominant anatomical modes. Psychiatric disorders typically exhibit weaker, more distributed effects requiring larger samples before multivariate models achieve higher accuracy.

\subsection{Relation to individual abnormality scores}

The same Mahalanobis distance provides a principled measure of how atypical an individual brain is relative to a healthy reference population:

\begin{equation}
M(\mathbf{x})^2
=
(\mathbf{x}-\boldsymbol{\mu}_0)^T
\Sigma^{-1}
(\mathbf{x}-\boldsymbol{\mu}_0).
\end{equation}

This connects classification and normative modeling. The same spectral structure that governs diagnostic accuracy also determines how precisely we can estimate centiles or deviation scores for individuals.

With small datasets, only the strongest modes are stable, limiting achievable AUC. As sample size grows, progressively weaker modes become identifiable, increasing Mahalanobis distance and improving classification accuracy.

Diagnostic performance improves as more disease-relevant spectral energy becomes estimable.

\subsection{Why the covariance spectrum is the bottleneck}

Diagnostic accuracy depends on the Mahalanobis distance

\begin{equation}
\Delta^2
=
\mathbf{d}^T
\Sigma^{-1}
\mathbf{d}.
\end{equation}

This expression makes it clear that classification improves as more modes become identifiable. However, the covariance matrix must itself be estimated from finite data.

At low sample size, estimates of $\Sigma$ can be unstable, especially in high dimensions. This instability affects the discriminant direction

\begin{equation}
\mathbf{w}
=
\Sigma^{-1}
\mathbf{d}.
\end{equation}

Sample covariance matrices obey concentration inequalities analogous to the DKW inequality:

\begin{equation}
\|\hat{\Sigma}-\Sigma\|
\sim
O\!\left(
\sqrt{\frac{p}{N}}
\right).
\end{equation}

Estimation error decreases slowly when the number of features $p$ is large.

The Davis-Kahan theorem shows that eigenvector accuracy depends on both the magnitude of perturbation and the spectral gap between neighboring eigenvalues \cite{daviskahan1970}:

\begin{equation}
\sin\theta_k
\le
C\,
\frac{\|\hat{\Sigma}-\Sigma\|}{\delta_k}.
\end{equation}

where $\delta_k$ is the gap between adjacent eigenvalues.

Eigenvectors are therefore stable only when eigenvalues are sufficiently separated.

This leads to the concept of \textbf{effective rank}. Even when the feature space has high dimension, only a smaller number of modes can typically be estimated reliably at a given sample size.

Steeper spectra concentrate meaningful variation into fewer dominant modes, which are easier to estimate reliably and contribute more strongly to Mahalanobis distance.

Representation learning methods often implicitly encourage low effective rank structure. Deep encoders transform the geometry of the data so disease-relevant variation becomes concentrated into a smaller number of stable directions.

\subsection{Sample size, Davis-Kahan, and the zeta law for AUC}

Combining Mahalanobis distance with Davis-Kahan bounds yields

\begin{equation}
\Delta^2(N)
\approx
\sum_{k=1}^{K(N)}
\frac{\alpha_k^2}{\lambda_k}.
\end{equation}

Sample covariance matrices satisfy concentration bounds

\begin{equation}
\|\hat{\Sigma}-\Sigma\|
\le
C
\sqrt{\frac{p}{N}}.
\end{equation}

Eigenvector stability requires

\begin{equation}
\sqrt{\frac{p}{N}}
\ll
\delta_k.
\end{equation}

Thus progressively weaker modes become identifiable as sample size increases.

Assuming power law decay,

\begin{equation}
\frac{\alpha_k^2}{\lambda_k}
\sim
k^{-\beta},
\end{equation}

we obtain

\begin{equation}
\Delta^2(N)
\sim
\sum_{k=1}^{K(N)}
k^{-\beta}.
\end{equation}

Because AUC depends monotonically on Mahalanobis distance,

\begin{equation}
AUC(N)
\approx
\Phi
\left(
\frac{1}{\sqrt{2}}
\left[
\sum_{k=1}^{K(N)}
k^{-\beta}
\right]^{1/2}
\right).
\end{equation}

\textbf{Diagnostic accuracy improves as sample size allows progressively weaker modes to become identifiable.}

As the number of identifiable modes grows with sample size, the cumulative signal approaches a truncated Riemann zeta function. In the limit of large data, \textbf{diagnostic accuracy approaches a form governed by the zeta function:}

\begin{tcolorbox}[colback=white,colframe=black,title=Zeta law for diagnostic accuracy]
\begin{equation}
AUC(N)
\approx
\Phi\!\left(
\frac{1}{\sqrt{2}}
\left[
\sum_{k=1}^{K(N)} k^{-\beta}
\right]^{1/2}
\right)
\rightarrow
\Phi\!\left(
\frac{1}{\sqrt{2}}
[\zeta(\beta)]^{1/2}
\right).
\end{equation}
\end{tcolorbox}

\subsection{Covariance concentration and identifiable modes}

For sub-Gaussian data, covariance concentration results give

\begin{equation}
\|\hat{\Sigma}-\Sigma\|_{\mathrm{op}}
\le
C\,\|\Sigma\|_{\mathrm{op}}
\left[
\sqrt{
\frac{r_{\mathrm{eff}}(\Sigma)+\log(1/\delta)}{N}
}
+
\frac{r_{\mathrm{eff}}(\Sigma)+\log(1/\delta)}{N}
\right],
\end{equation}

where

\begin{equation}
r_{\mathrm{eff}}(\Sigma)
=
\frac{\operatorname{tr}(\Sigma)}
{\|\Sigma\|_{\mathrm{op}}}.
\end{equation}

For large $N$,

\begin{equation}
\|\hat{\Sigma}-\Sigma\|_{\mathrm{op}}
\approx
C
\|\Sigma\|_{\mathrm{op}}
\sqrt{
\frac{r_{\mathrm{eff}}+\log(1/\delta)}{N}
}.
\end{equation}

Eigenvalues satisfy Weyl's inequality:

\begin{equation}
|\hat{\lambda}_k-\lambda_k|
\le
\|\hat{\Sigma}-\Sigma\|_{\mathrm{op}}.
\end{equation}

Eigenvector accuracy follows from Davis-Kahan:

\begin{equation}
\sin\theta_k
\le
C\,
\frac{
\|\hat{\Sigma}-\Sigma\|_{\mathrm{op}}
}{\Delta_k}.
\end{equation}

and a mode is identifiable when

\begin{equation}
\|\hat{\Sigma}-\Sigma\|_{\mathrm{op}}
\ll
\Delta_k.
\end{equation}

\subsection{Power law eigenspectra}

If eigenvalues follow a power law,

\begin{equation}
\lambda_k
\sim
k^{-\gamma},
\end{equation}

then spectral gaps behave approximately as

\begin{equation}
\Delta_k
\sim
k^{-(\gamma+1)}.
\end{equation}

Combining this with covariance concentration gives

\begin{equation}
K(N)
\sim
N^{\frac{1}{2(\gamma+1)}}.
\end{equation}

Thus \textbf{the number of reliable modes grows predictably with sample size.}

This gives a simple scaling-law view of sample complexity: \textbf{to reach a target level of accuracy, the required sample size depends on how quickly the truncated zeta sum approaches its limit:}

\begin{tcolorbox}[colback=white,colframe=black,title=Zeta law for sample complexity]
\begin{equation}
\Delta^2(N)
\approx
C_d
\sum_{k=1}^{K(N)} k^{-\beta},
\qquad
K(N)\sim N^{\frac{1}{2(\gamma+1)}}.
\end{equation}

\begin{equation}
\Delta^2(N)
\approx
C_d\,H_{K(N)}^{(\beta)}
\;\longrightarrow\;
C_d\,\zeta(\beta),
\qquad
AUC(N)\to AUC_{\infty}.
\end{equation}
\end{tcolorbox}

\subsection{Compact zeta law expression}

Combining results gives

\begin{equation}
\Delta^2(N)
\approx
C_d
\sum_{k=1}^{K(N)}
k^{-\beta}.
\end{equation}

Hence

\begin{equation}
AUC(N)
\approx
\Phi
\left(
\frac{1}{\sqrt{2}}
\left[
C_d
H_{K(N)}^{(\beta)}
\right]^{1/2}
\right).
\end{equation}

For $\beta>1$,

\begin{equation}
AUC_\infty
\approx
\Phi
\left(
\frac{1}{\sqrt{2}}
\left[
C_d
\zeta(\beta)
\right]^{1/2}
\right).
\end{equation}

This expression links \textbf{predictive accuracy to sample size, effective rank, covariance spectral slope, and disease-aligned spectral decay.}

\begin{figure}[ht]
\centering
\includegraphics[width=0.8\linewidth]{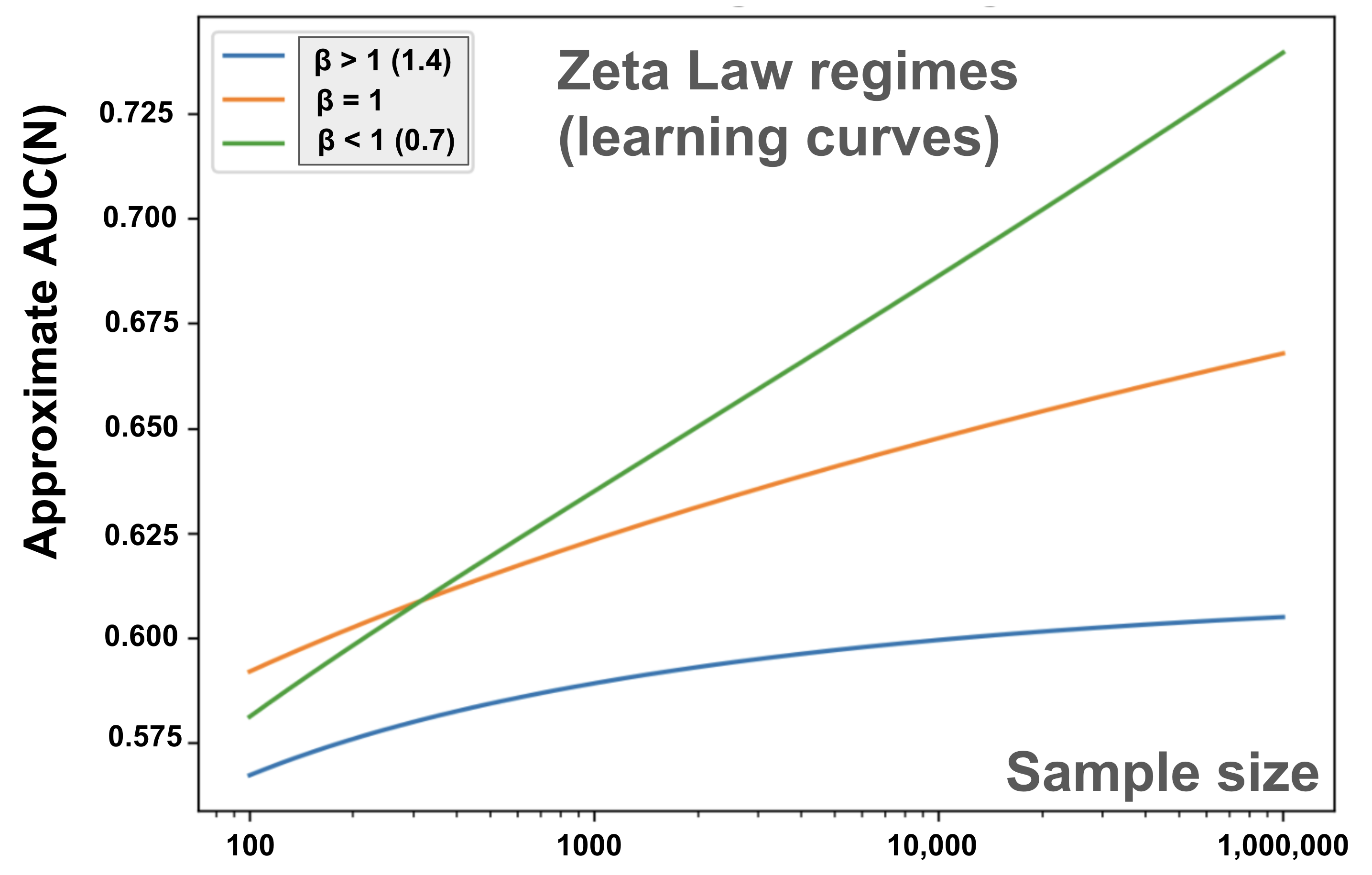}

\caption{
\textbf{Will Discovery be Fast or Slow?} Learning curves predicted by the zeta law under different spectral decay rates. When signal is concentrated in a small number of stable modes, accuracy improves rapidly with sample size, whereas diffuse signals require substantially more data before meaningful gains appear.
}

\label{fig:3curves}
\end{figure}

\subsection{The Tower of Hanoi}

The truncated power-law sum
\begin{equation}
\Delta^2(N)
\;\propto\;
\sum_{k=1}^{K(N)} k^{-\beta}
\end{equation}

can be interpreted as the cumulative signal energy contributed by progressively weaker eigenmodes as sample size increases. Each spectral mode contributes a decreasing amount of signal energy, and only modes whose eigenvalues can be estimated reliably contribute to classification accuracy.

\begin{figure}[t]
\centering
\includegraphics[width=0.95\linewidth]{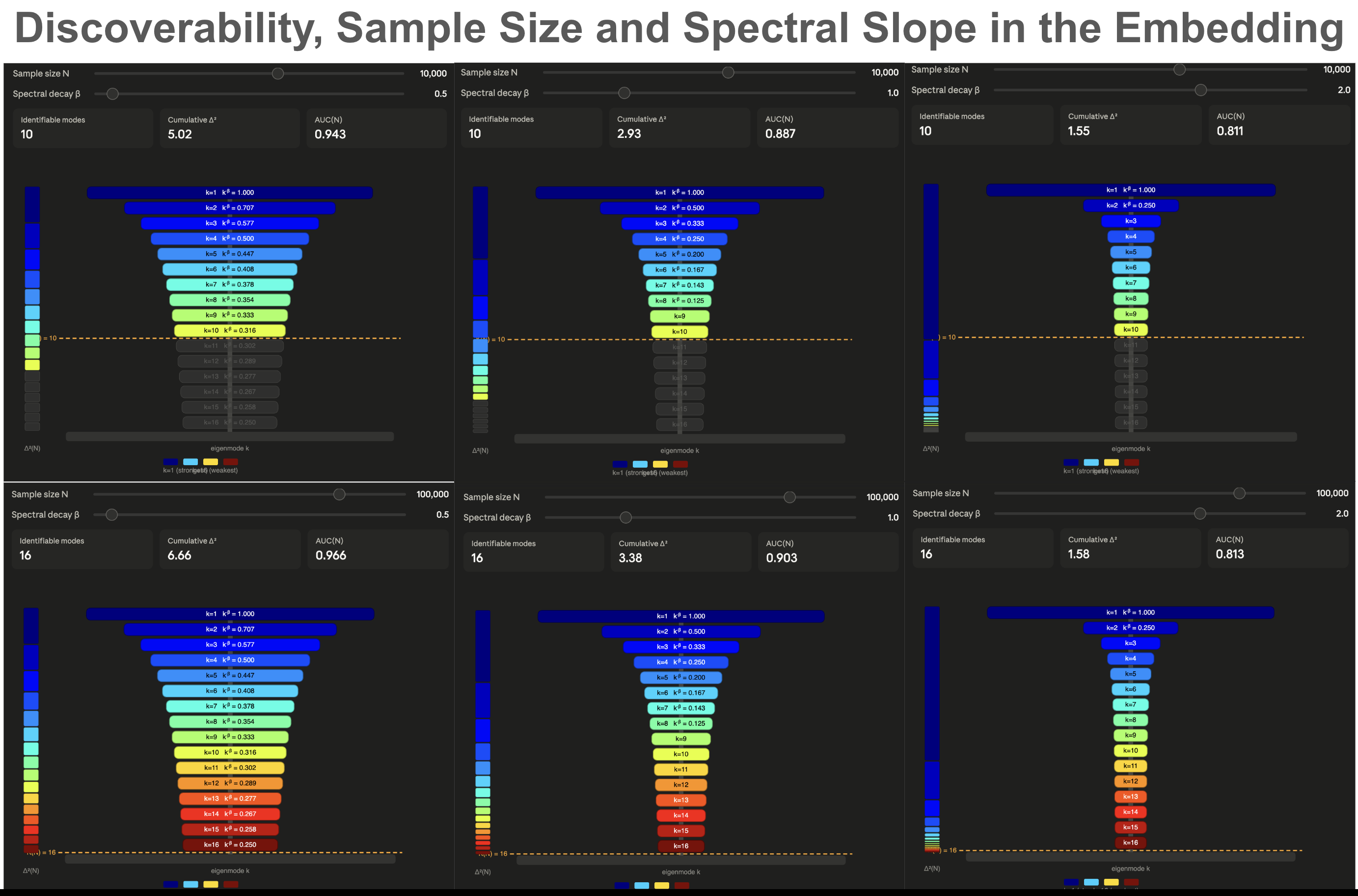}

\caption{
\textbf{Tower of Hanoi view of partial zeta sums governing discoverability.}
Each column shows cumulative Mahalanobis signal 
$\Delta^2(N) \propto \sum_{k=1}^{K(N)} k^{-\beta}$ 
as progressively weaker eigenmodes become identifiable. 
Colored disks represent spectral modes ordered by strength, with disk size reflecting signal contribution and the dashed line indicating the identifiability threshold $K(N)$. 
Flatter spectra ($\beta=0.5$, left) distribute signal across many weak modes, so cumulative signal grows slowly with increasing sample size. 
Intermediate decay ($\beta=1$, center) yields gradual accumulation across modes. 
Steeper spectra ($\beta=2$, right) concentrate signal in a few dominant modes, so most detectable signal is captured even at smaller sample sizes. 
The bottom row shows how increasing sample size from $N=10{,}000$ to $100{,}000$ increases the number of identifiable modes and raises cumulative signal, with the rate of gain governed by the spectral slope $\beta$.
}

\label{fig:hanoi}
\end{figure}

Figure~\ref{fig:hanoi} visualizes this accumulation of signal using a Tower-of-Hanoi analogy (Tower of Hanoi is a children's toy, based on a stack of colored rings of different sizes). Each disk represents an eigenmode, ordered by decreasing signal strength. Disk size corresponds to the magnitude of spectral contribution $k^{-\beta}$, and the dashed line indicates the identifiability threshold $K(N)$ determined by sample size. Only disks above this threshold contribute reliably to predictive accuracy.

When the spectral decay is shallow (small $\beta$), signal is distributed across many weak modes, so cumulative signal grows slowly as additional modes become identifiable. When spectral decay is steeper (larger $\beta$), signal is concentrated in fewer dominant modes, so most useful signal can be captured with fewer samples. Increasing sample size increases $K(N)$, allowing progressively weaker modes to contribute to the Mahalanobis distance and improving classification performance.

This visualization highlights a central idea of the zeta law: the efficiency with which predictive signal accumulates depends on how rapidly spectral energy decays across identifiable modes.

\begin{table}[h]
\centering
\begin{tabular}{rrrrrr}
\toprule
$N$ & $\beta$ & $K(N)$ & $\Delta^2$ & $\mathrm{AUC}(N)$ \\
\midrule
10,000  & 0.5 & 10 & 5.02 & 0.943 \\
10,000  & 1.0 & 10 & 2.93 & 0.887 \\
10,000  & 2.0 & 10 & 1.55 & 0.811 \\
100,000 & 0.5 & 18 & 7.14 & 0.971 \\
100,000 & 1.0 & 18 & 3.50 & 0.907 \\
100,000 & 2.0 & 18 & 1.59 & 0.814 \\
\bottomrule
\end{tabular}
\caption{Identifiable modes $K(N)$, cumulative Mahalanobis signal $\Delta^2(N)$, and diagnostic accuracy $\mathrm{AUC}(N)$ for each panel of Figure~\ref{fig:hanoi}, across sample sizes and spectral decay rates.}
\label{tab:hanoi}
\end{table}

\subsection{Regimes of discovery}

The exponent $\beta$ determines how rapidly signal accumulates as additional modes become identifiable.

When $\beta>1$, most signal lies in a few dominant modes and discovery is fast. But if $\beta \approx 1$, signal is distributed across many modes and progress is more gradual. And when $\beta<1$, signal is very diffuse and it may help more to improve representations than just carrying on increasing the sample size.

\subsection{Ways to improve discovery}

The zeta law suggests several ways to improve predictive performance, such as (1) increasing the \textbf{sample size} to reveal weaker modes; (2) improving the \textbf{encoders} to concentrate signal into earlier modes; (3) designing informative \textbf{features} that align disease effects with dominant modes;
(4) integrating multiple modalities to increase \textbf{shared spectral energy}; (5) reducing nuisance variation through \textbf{harmonization or noise modeling}. Different problems fall into distinct spectral categories depending on how rapidly signal is decaying across modes. When signal is concentrated, larger datasets may be enough. For distributed signals, better representations may help, and if signal is diffuse, multimodal alignment or improved feature design may help the most.

\section{Improving Discoverability with Better Encoders}

Different models alter the spectrum of the data representation. Successful models concentrate disease-relevant variation into fewer stable modes, effectively increasing $\beta$ and reducing the sample size required for reliable prediction.

Below we summarize how commonly used statistical and deep learning models affect spectral structure.

\subsection{Linear discriminant analysis (baseline spectrum)}

Linear discriminant analysis defines the optimal linear classifier

\begin{equation}
\mathbf{w} \propto \Sigma^{-1}\mathbf{d}
\end{equation}

yielding the Mahalanobis distance

\begin{equation}
\Delta^2
=
\sum_k
\frac{\alpha_k^2}{\lambda_k}.
\end{equation}

LDA does not alter the spectrum, but reveals the inherent decay of disease-aligned energy across modes. The rate of discovery is therefore determined by $\beta$.

\subsection{Elastic net (spectral shrinkage)}

Elastic net stabilizes estimation by shrinking unstable directions:

\begin{equation}
\hat{\mathbf{w}}
=
\arg\min_{\mathbf{w}}
\|\mathbf{y}-X\mathbf{w}\|^2
+
\lambda_1\|\mathbf{w}\|_1
+
\lambda_2\|\mathbf{w}\|_2^2.
\end{equation}

Ridge shrinkage increases small eigenvalues:

\begin{equation}
\lambda_k
\rightarrow
\lambda_k + \lambda_2.
\end{equation}

This reduces sensitivity to noise and increases spectral gaps. Sparsity suppresses weak unstable modes, effectively steepening the decay of

\begin{equation}
\frac{\alpha_k^2}{\lambda_k}.
\end{equation}

Elastic net therefore increases $\beta$ and improves sample efficiency.

\subsection{Latent variable models and dVAEs (low-rank compression)}

Latent encoders such as variational autoencoders transform the covariance structure of the input data:

\begin{equation}
\Sigma_z
\approx
J_f
\Sigma_x
J_f^T.
\end{equation}

When nuisance variation is compressed more strongly than disease-relevant variation, the latent eigenspectrum decays more rapidly:

\begin{equation}
\lambda_k^{(z)}
\sim
k^{-\gamma'},
\qquad
\gamma' > \gamma.
\end{equation}

KL regularization encourages compact representations, packing useful signals into fewer stable modes and increasing effective $\beta$.

\subsection{CCA (cross-modal spectral concentration)}

Canonical correlation analysis identifies directions maximizing cross-modal correlation:

\begin{equation}
\max_{\mathbf{u},\mathbf{v}}
\mathbf{u}^T
\Sigma_{xy}
\mathbf{v}.
\end{equation}

Recent work has shown that stability of CCA and PLS solutions depends strongly on spectral decay and sample size \cite{helmer2024cca}. Leading singular values capture shared biological structure:

\begin{equation}
\sigma_k(\Sigma_{xy})
\sim
k^{-\beta_{\mathrm{cross}}}.
\end{equation}

CCA concentrates shared signals into early modes, increasing identifiable spectral energy.

This principle underlies many modern multimodal representation learning approaches, including contrastive learning methods that align images and text into a shared embedding space.

\subsection{Vision-language models (alignment increases early singular values)}

Contrastive pretraining objectives maximize similarity between paired embeddings:

\begin{equation}
\mathcal{L}
=
-
\log
\frac{
\exp(\mathbf{z}_i^T\mathbf{t}_i/\tau)
}{
\sum_j
\exp(\mathbf{z}_i^T\mathbf{t}_j/\tau)
}.
\end{equation}

Such objectives increase dominant singular values of the cross-modal operator. Even weak modalities can improve classification when alignment concentrates shared information into fewer modes, effectively increasing $\beta$ in the joint representation. Vision-language models embed images and text into a shared vector space, often with dimensionality on the order of hundreds. This allows information from different modalities to reinforce shared latent structure.

\subsection{Transformers (adaptive spectral filtering)}

Self-attention defines a learned kernel:

\begin{equation}
\mathrm{Attn}(X)
=
\mathrm{softmax}
\left(
\frac{QK^T}{\sqrt{d}}
\right)
V.
\end{equation}

Self-attention adaptively emphasizes directions with strong shared covariance. This produces feature spaces with steeper effective spectra than raw inputs, analogous to kernel CCA. Transformers compute relationships between tokens using learned projections of queries and keys. Vision transformers apply this principle to image patches, but the underlying mechanism is similar across modalities.

\subsection{Summary}

Across these models, there is a common pattern. We get improved discoverability when new representations increase the early spectral energy or enlarge spectral gaps.

In terms of the zeta law, successful models increase $\beta$ by concentrating disease-relevant variation into earlier modes, reducing the sample size required for reliable prediction.

\section{Thought Experiments and Applications of the Zeta Law}

\subsection{Multi-disease classification and required sample size}

We began with the question of how much data is needed to classify multiple brain disorders with a target level of accuracy. Suppose we want to distinguish among $D$ diseases such as Alzheimer's disease, Parkinson's disease, schizophrenia, and autism. Given a feature representation or embedding, each disease corresponds to a direction in the feature space describing the contrast between cases and controls.

Under the zeta law, the performance of a disease classifier depends on Mahalanobis energy in whitened space:

\begin{equation}
E_d
=
\mathbf{d}^T \Sigma^{-1} \mathbf{d}
=
\sum_{k=1}^{K(N)}
\frac{(\mathbf{u}_k^T \mathbf{d})^2}{\lambda_k}.
\end{equation}

Only the first $K(N)$ eigenmodes can be estimated reliably at sample size $N$. If diseases are modeled independently, the required sample size is determined by the most difficult disease to classify:

\begin{equation}
N_{\mathrm{required}}
\propto
\max_j
\frac{1}{E_{d_j}}.
\end{equation}

In this setting, the weakest signal determines study size even if other diseases are easier to classify.

However, many disorders share partially overlapping spatial patterns, suggesting disease effects lie in a low-dimensional subspace of brain organization. For example, our ENIGMA Consoritum has shown that cortical thickness alterations across multiple disorders can often be summarized by a small number of principal axes \cite{hettwer2022transdiagnostic,hettwer2026axes}, implying strong alignment of disease contrast vectors.

In the zeta framework, this indicates low effective rank of the disease operator, concentrating signal energy into leading eigenmodes. As the number of stable modes grows approximately as

\begin{equation}
K(N)
\sim
N^{\frac{1}{2(\gamma+1)}},
\end{equation}

low-rank disease structure reduces the sample size required to estimate relevant modes and achieve a target AUC.

If disease contrasts lie in a shared subspace of rank $r \ll D$, joint estimation can increase signal captured in leading eigenmodes. Methods such as CCA or multitask representation learning estimate projections maximizing covariance between imaging features $\mathbf{X}$ and disease labels $\mathbf{Y}$:

\begin{equation}
\max_{\mathbf{w}_x,\mathbf{w}_y}
\mathrm{corr}(\mathbf{X}\mathbf{w}_x,\mathbf{Y}\mathbf{w}_y).
\end{equation}

Concentrating signal into fewer eigenmodes steepens the effective spectrum, increasing recoverable Mahalanobis energy at fixed $N$. The number of stable modes grows approximately according to:

\begin{equation}
K(N)
\sim
N^{1/\beta}.
\end{equation}

In the ideal case of strongly aligned disease effects,

\begin{equation}
N_{\mathrm{joint}}
\approx
N_{\mathrm{single}}
\cdot
\frac{r}{D},
\end{equation}

suggesting large gains when diseases share low-rank structure.

The zeta law predicts three situations we can be in: (1) independent diseases, where sample size is governed by the hardest disease; (2) partially shared structure, where joint models reduce required $N$; and (3) strongly shared structure, where many diseases can be classified efficiently in a common latent space. Practically, this suggests estimating a shared disease subspace before fitting disease-specific classifiers.

\subsection{Relation to anatomical gradients}

Suppose we represent each disease by a vector of standardized effect sizes across brain features and define the disease matrix

\begin{equation}
D
=
[\mathbf{d}_1,\mathbf{d}_2,\dots,\mathbf{d}_D]
\in
\mathbb{R}^{p \times D}.
\end{equation}

The matrix

\begin{equation}
D D^T
=
\sum_{j=1}^{D}
\mathbf{d}_j \mathbf{d}_j^T
\end{equation}

summarizes how diseases align in brain space. Eigenvectors correspond to shared anatomical gradients and eigenvalues indicate how strongly diseases concentrate along those gradients. If most variance is captured by a few eigenvalues, fewer modes must be estimated reliably. Under the zeta law,

\begin{equation}
K(N)
\sim
N^{1/\beta},
\end{equation}

joint modeling of related diseases reduces the sample size needed to achieve a target AUC.

\subsection{Discovering genetic effects on brain structure}

In imaging genetics, the goal is often to explain variance in brain measures using genomic data. Genome-wide association studies show that the spectrum of genetic effects is often very flat, with many variants having small effects. Let genetic variation be represented by $\mathbf{X}$ and imaging features by $\mathbf{Y}$. The cross-modal covariance operator

\begin{equation}
C_{XY}
=
\frac{1}{N}
\mathbf{X}^T
\mathbf{Y}
\end{equation}

summarizes how genetic variation predicts brain variation. Its singular value decomposition

\begin{equation}
C_{XY}
=
U \Sigma V^T
\end{equation}

reveals coupled modes of variation between genome and brain.

When singular values decay slowly, the spectrum is flat and high effective dimensionality implies large sample sizes are required.

Under the zeta law,

\begin{equation}
K(N)
\sim
N^{\frac{1}{2(\gamma+1)}},
\end{equation}

so flat spectra imply very large datasets are needed before substantial variance can be explained.

Polygenic risk scores can be interpreted as projections of genotype data onto directions aligned with phenotype variation.

Aggregating many weak SNP effects into composite variables concentrates signal energy into earlier modes, steepening the spectrum of $C_{XY}$ and improving sample efficiency.

More generally, representation learning methods aim to learn encoders

\begin{equation}
\mathbf{z}_x = f(\mathbf{X}),
\qquad
\mathbf{z}_y = g(\mathbf{Y})
\end{equation}

such that cross-modal covariance

\begin{equation}
\mathrm{cov}(\mathbf{z}_x,\mathbf{z}_y)
\end{equation}

is concentrated in a small number of dominant singular values.

CCA, PLS, and contrastive learning methods can be interpreted as learning projections aligning leading eigenmodes of two modalities.

\subsection{The three-modality paradox}

An interesting phenomenon arises when adding a third modality such as text descriptions, clinical covariates, or ontology terms. Vision-language models can perform as well as or better than purely image-based classifiers even when text provides limited additional predictive information. The zeta framework suggests the benefit arises because the additional modality may have lower rank or steeper spectrum, providing a stable coordinate system aligning signals across modalities.

Let image features $\mathbf{X}$, disease labels $\mathbf{Y}$, and text embeddings $\mathbf{T}$ define three representations. Pairwise cross-modal operators can be written as

\begin{equation}
C_{XY} = \mathbf{X}^T \mathbf{Y},
\qquad
C_{XT} = \mathbf{X}^T \mathbf{T},
\qquad
C_{YT} = \mathbf{Y}^T \mathbf{T}.
\end{equation}

Even if $\mathbf{T}$ contains limited predictive information, it may concentrate shared structure into fewer dominant eigenmodes.

When one modality has a steeper spectrum, alignment with that modality can increase the proportion of signal energy captured in early modes of the other modalities. In this case, the third modality acts as a spectral regularizer. As the number of reliably estimable modes grows approximately as

\begin{equation}
K(N)
\sim
N^{\frac{1}{2(\gamma+1)}},
\end{equation}

alignment to a lower-rank modality increases effective $\beta$ and lowers the sample size needed for accurate prediction.

Pairwise contrastive learning objectives can therefore improve classification even when auxiliary modalities contain limited direct predictive information. An additional advantage is robustness to missing data. Pairwise alignment allows each modality to connect through shared latent structure without requiring complete observations across all modalities.

\subsection{A note on joint multimodal objectives}

Fully joint alignment of three modalities can be achieved with objectives that maximize agreement across triplets of embeddings, such as \emph{Symile} \cite{saporta2024symile}. A simplified objective can be written as

\begin{equation}
\mathcal{L}_{\mathrm{symile}}
=
-
\log
\frac{
\exp(\langle \mathbf{z}_x,\mathbf{z}_y,\mathbf{z}_t \rangle)
}{
\sum_{t'}
\exp(\langle \mathbf{z}_x,\mathbf{z}_y,\mathbf{z}_{t'} \rangle)
}.
\end{equation}

Such formulations appear in multimodal transformers using higher-order attention mechanisms. In biomedical settings, modalities often contain both shared and modality-specific variation. Pairwise encoders may therefore offer advantages, allowing each modality pair to align where shared signal exists while preserving modality-specific structure elsewhere. As sample size increases, the dimension of the reliably estimable shared subspace is expected to grow because additional weak cross-modal associations become detectable.

Under the zeta law,

\begin{equation}
K(N)
\sim
N^{1/\beta},
\end{equation}

so progressively finer cross-modal structure becomes estimable. Higher-order multimodal objectives may  capture richer shared structure in the end, but pairwise objectives may be more sample efficient at current dataset sizes.

\section{Improving AUC(N) with Better Features: Lifts, Kernels, and HSIC}

\subsection{Why feature design helps in very high dimensions}

Many biomedical data types, such as resting state functional MRI and genomics, are extremely high dimensional. In practice, researchers rarely input raw measurements directly into predictive models. Instead, derived representations are constructed such as correlations, gradients, ICA components, or connectivity summaries. These transformations can be understood as feature lifts that increase statistical dependence between representation and phenotype, concentrating signal energy into fewer eigenmodes. Under the zeta law, predictive accuracy depends on how quickly signal accumulates in stable modes. Raw voxelwise or time-series representations often have relatively flat spectra, meaning signal is spread across many weak directions.

Carefully designed features can steepen spectral decay, increasing $\beta$ and allowing predictive structure to be estimated reliably at smaller sample sizes.

\subsection{Feature lifts increase dependence structure}

A \emph{feature lift} maps data into a representation in which relationships among variables become more informative for prediction. Instead of using raw measurements $\mathbf{x}$, we construct transformed features

\begin{equation}
\boldsymbol{\phi}(\mathbf{x})
\end{equation}

designed to increase statistical dependence with the outcome.

Pearson correlations in resting state functional MRI capture relationships between regions rather than marginal activity levels. Asymmetry measures capture differences between left and right brain regions. Many disease effects are easier to detect by creating a model of their relational structure. In spectral terms, lifts increase alignment between representation and phenotype, concentrating signal energy into leading eigenmodes and improving statistical efficiency.

\subsection{Why functional MRI models rarely operate on raw time series}

Deep learning models applied to resting state functional MRI of the brain rarely operate directly on voxelwise time series even though this would allow highly flexible modeling. Most pipelines construct intermediate representations such as functional connectivity matrices, ICA components, graph summaries, or gradient embeddings. Empirically, disease-related signals often appear as distributed network structure rather than localized voxel effects. From the zeta law perspective, these transformations increase alignment between signal and stable eigenmodes, reduce effective rank of the disease operator, and increase $\beta$. The widespread use of connectivity features therefore reflects an implicit attempt to improve scaling of AUC with sample size.

\subsection{Examples of useful lifts}

Many widely used neuroimaging features can be interpreted as lifts that improve spectral geometry. Connectivity matrices express pairwise interactions between regions, and gradient representations reveal large-scale axes of cortical organization \cite{margulies2016situating}.
Topological data analysis is an example of a higher-order lift. Persistent homology - recently applied to brain imaging by the ENIGMA-OCD international working group \cite{ruan2026hodge} - summarizes loops and higher-order structure in connectivity networks, capturing organizational properties not represented in pairwise correlations. Such representations show how creative feature construction can reveal stable structure that improves statistical efficiency.

\subsection{Lifts as nonlinear feature maps}

Many lifts can be interpreted as nonlinear mappings into richer feature spaces,

\begin{equation}
\boldsymbol{\phi}: X \rightarrow \mathcal{H}
\end{equation}

where relationships between variables become more linearly structured. Kernel methods, neural network embeddings, graph features, and topological summaries all construct such mappings. In the zeta framework, useful lifts steepen spectral decay of the cross-modal operator, increasing the proportion of signal energy captured in early modes. Improved features therefore increase statistical dependence between representation and phenotype, improving sample efficiency.

\subsection{Hilbert-Schmidt Independence Criterion (HSIC)}

The Hilbert-Schmidt Independence Criterion provides a formal measure of dependence between representations and outcomes in lifted feature spaces.

Let kernels $k(x,x')$ and $l(y,y')$ define similarity functions in input and output spaces. HSIC measures the squared Hilbert-Schmidt norm of the cross-covariance operator:

\begin{equation}
\mathrm{HSIC}(X,Y)
=
\|C_{XY}\|_{HS}^2
\end{equation}

where

\begin{equation}
C_{XY}
=
\mathbb{E}
\left[
(\boldsymbol{\phi}(X)-\boldsymbol{\mu}_X)
\otimes
(\boldsymbol{\psi}(Y)-\boldsymbol{\mu}_Y)
\right].
\end{equation}

Equivalently,

\begin{equation}
\mathrm{HSIC}(X,Y)
=
\sum_k
\sigma_k^2,
\end{equation}

where $\sigma_k$ are singular values of the cross-modal operator.

Lifts that increase HSIC concentrate signal energy into dominant modes. This formalizes the intuition underlying feature engineering. Useful lifts alter geometry so informative structure appears earlier in the spectrum.

\subsection{Relationship among HSIC, CCA, and contrastive learning}

HSIC, CCA, and contrastive learning are closely related approaches to increase statistical dependence between representation and target. HSIC measures total dependence:

\begin{equation}
\mathrm{HSIC}(X,Y)
=
\|C_{XY}\|_{HS}^2.
\end{equation}

CCA identifies directions maximizing normalized covariance:

\begin{equation}
\max_{\mathbf{u},\mathbf{v}}
\mathrm{corr}
(
\mathbf{u}^T \boldsymbol{\phi}(X),
\mathbf{v}^T \boldsymbol{\psi}(Y)
).
\end{equation}

Contrastive learning objectives such as CLIP \cite{radford2021clip} encourage shared modes to dominate nuisance variation:

\begin{equation}
\mathcal{L}_{\mathrm{CLIP}}
=
-\sum_{i=1}^N
\log
\frac{
\exp(
\langle
\mathbf{z}_i^x,
\mathbf{z}_i^y
\rangle
/\tau
)
}{
\sum_{j=1}^N
\exp(
\langle
\mathbf{z}_i^x,
\mathbf{z}_j^y
\rangle
/\tau
)
}.
\end{equation}

All three approaches alter the spectrum of the shared operator so informative structure appears earlier. Representations producing stronger leading modes and faster spectral decay are expected to yield more data-efficient predictions.

\section{Conclusion: When do deep models outperform linear models?}

A key question in this whole paper has been when complex models such as deep neural networks outperform simpler linear approaches. The zeta law suggests that performance depends on how efficiently signal energy concentrates into stable modes that can be estimated from finite data. The key quantity is spectral decay of the representation. If signal energy decays slowly across modes, many weak directions must be estimated before predictive structure becomes detectable. In this regime, simpler models often perform best because they focus estimation on strongest modes. But if spectral decay is steeper, more signal is captured in early modes, allowing higher-capacity models to exploit additional structure as sample size increases. This leads to characteristic cross-over behavior where linear or low-rank models perform well at smaller sample sizes, but nonlinear or higher-capacity models eventually outperform them as more modes can eventually be estimated (see cross-over figure).

\begin{figure}[ht]
\centering
\includegraphics[width=0.8\linewidth]{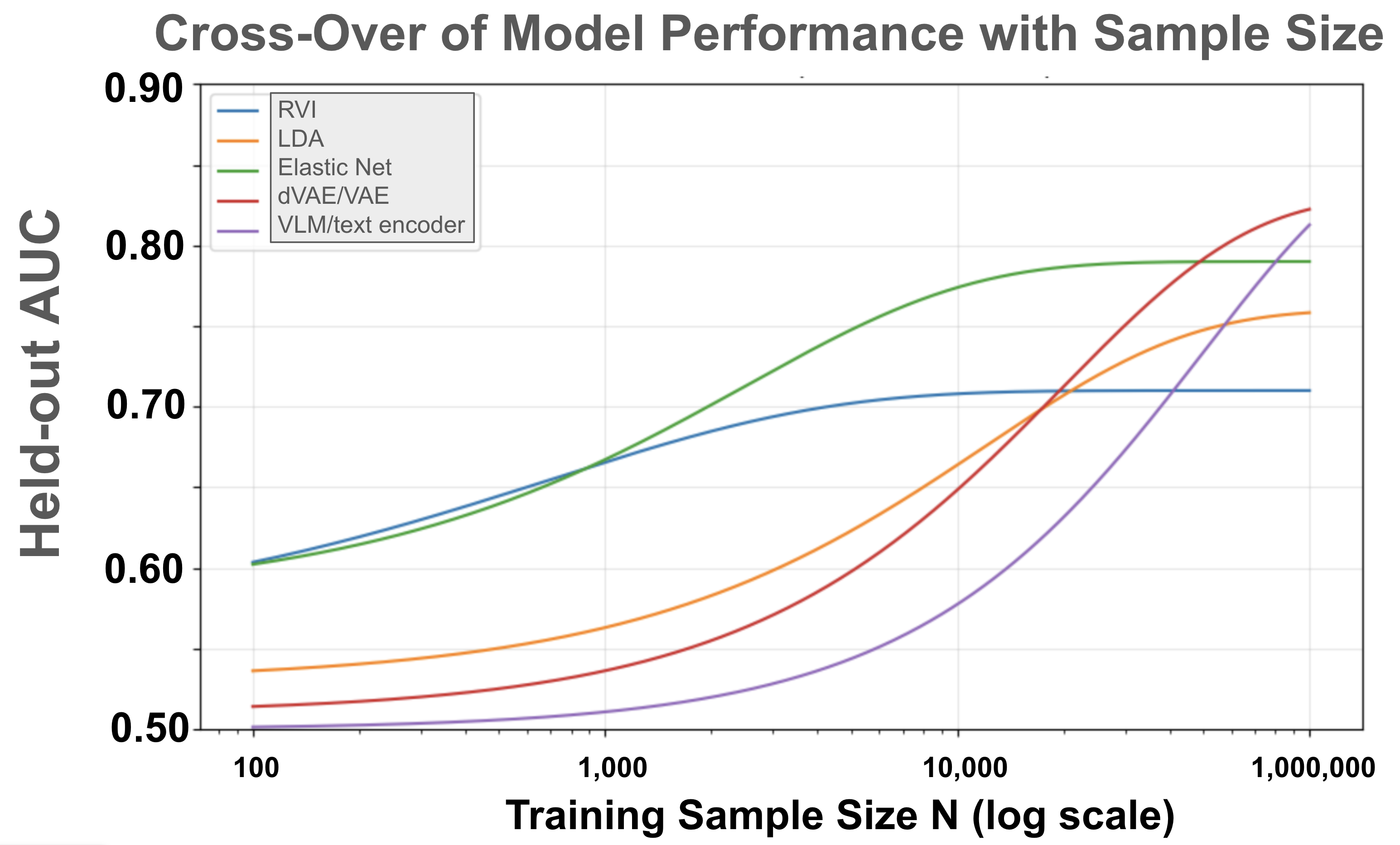}

\caption{
\textbf{Expected cross-over ordering of model performance under the zeta law. Cross-over occurs when sample size is sufficient to estimate the useful degrees of freedom of richer representations}. As $N$ increases, performance may progress from LDA (with RVI as a covariance-agnostic approximation), to elastic net, to dVAE, with a VLM plus auxiliary text encoder crossing over last but potentially reaching the highest asymptotic $AUC(N)$. Simpler linear models perform best at small $N$ when disease effects are smooth or low rank, whereas higher-capacity models benefit once sufficient data exist to estimate sparse, nonlinear, or cross-modal structure that concentrates signal into earlier spectral modes.
}

\label{fig:crossover}
\end{figure}

The cross-over point depends on spectral decay of the encoder and strength of underlying signal. At small sample sizes, linear models may perform best because they estimate only strongest directions. As sample size increases, richer representations such as deep encoders, multimodal embeddings, or lifted feature spaces can use additional structure and increase AUC more rapidly.

\subsection{Implications for neuroimaging research}

Different brain imaging modalities have different spectral properties, which may help us to anticipate how different models will perform. Structural and diffusion MRI show moderately structured variation aligned with developmental and disease gradients, suggesting intermediate spectral decay. By contrast, resting-state functional MRI, which consists of 4D videos of brain activity, is extremely high dimensional and, without further feature extraction, it can exhibit relatively flat spectra when represented at voxel or time-series level. Encoders such as connectivity matrices, gradients, or tokenized brain representations can steepen spectral decay by concentrating signal into more stable modes.
With this level of complexity in the inputs, carefully designed feature representations have a strong chance of outperforming the raw high-dimensional inputs.

\subsection{Practical tactics across spectral regimes}

Different spectral situations suggest different modeling strategies. When spectra are relatively flat (small $\alpha$), signal is distributed across many weak modes. Models that constrain effective dimensionality tend to perform best in this setting. Low-rank models, strong regularization, and carefully designed feature lifts can improve efficiency by concentrating signal into earlier identifiable modes. Multimodal alignment, kernel methods, and dependence-maximizing representations may also help. At intermediate spectral decay rates, signal is partly concentrated in early modes but higher modes still contain useful information. Multitask learning, nonlinear feature lifts, CCA, or contrastive learning may help capture shared structure without requiring extremely large sample sizes. Moderately expressive neural networks may also perform well when guided by representations that already concentrate meaningful variation.

When spectra are steeper (larger $\alpha$), signal is concentrated in relatively few stable modes. Higher-capacity models can exploit richer structure as sample size increases. In this regime, deep neural networks, multimodal transformers, and flexible nonlinear models may achieve higher accuracy because sufficient data exist to estimate higher-order interactions reliably.

Across these regimes, we need to match representation and model complexity to the spectral structure of the problem. Methods that increase dependence structure, align shared subspaces, or reduce effective dimensionality can shift problems toward regimes in which richer models become advantageous at smaller sample sizes.

\begin{figure}[ht]
\centering
\includegraphics[width=0.8\linewidth]{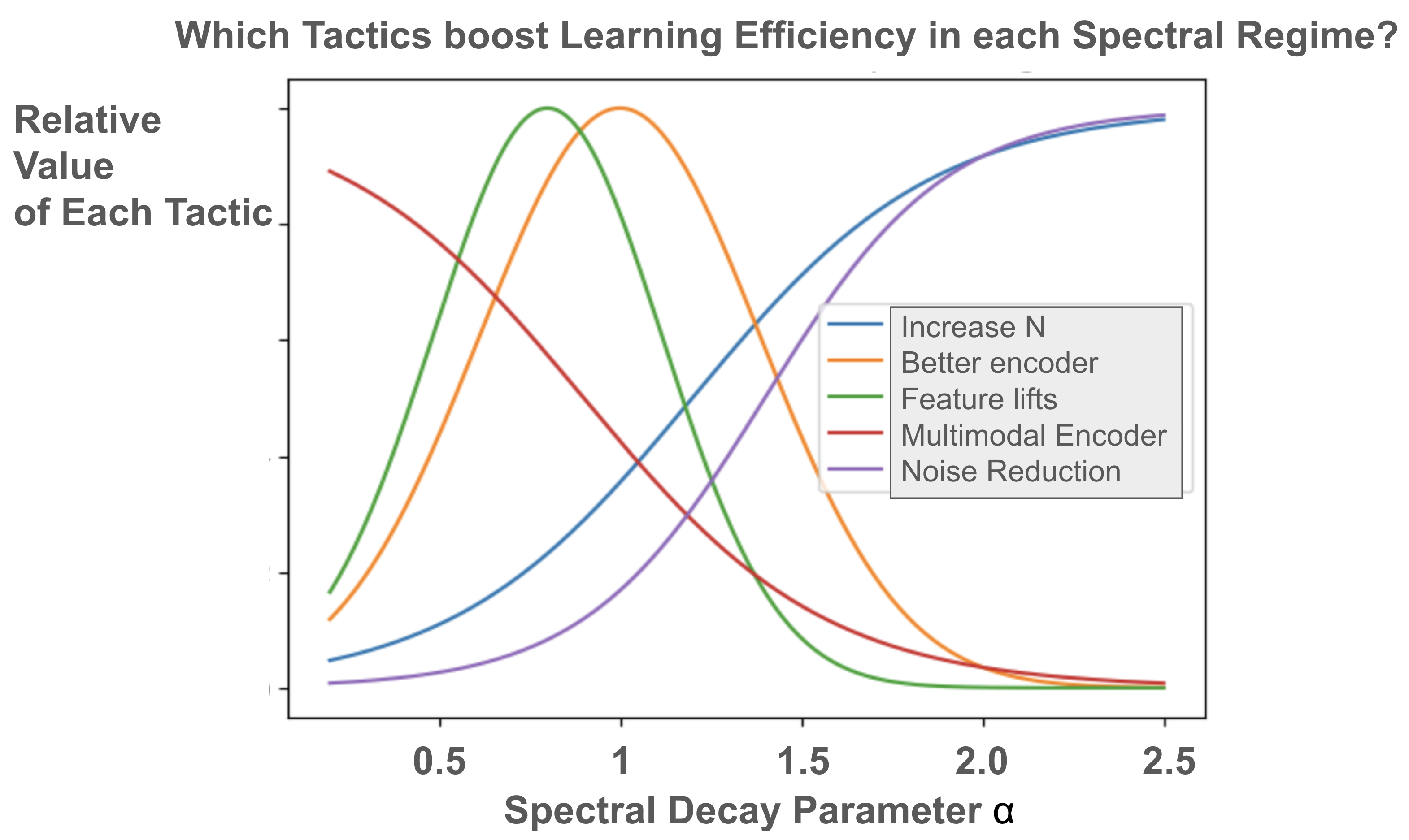}
\caption{
\textbf{Different paths to better prediction accuracy.} The best strategy depends on how widely the disease signal is spread across patterns of variation. If signal is diffuse, combining multiple data types may help reveal shared structure and strengthen weak effects. As structure becomes clearer, cleverly designed features such as connectivity, gradients, asymmetry, or topological summaries may start to identify relationships that are easier to detect. With more structured signals, improved encoders or pretrained models can capture useful nonlinear patterns that simpler approaches may miss. If signal is already concentrated in stable patterns, collecting more data becomes especially valuable because additional structure can be reliably estimated. When signals are strong but noisy, reducing measurement noise can boost accuracy. Overall, prediction accuracy may increase most rapidly when analytic choices match how biological signal is organized across patterns of variation.
}
\label{fig:levers}
\end{figure}

\subsection{Interpretation}

The zeta law predicts that progress in making biomedical discoveries should come from increasing sample size, improving representations to steepen spectral decay, and aligning modalities or tasks to increase shared low-rank structure. These strategies should all shift the cross-over point at which richer models perform best.

By increasing sample size, we should be able to estimate  progressively weaker modes reliably. When spectra are flatter, this process is slow because signal is distributed across many weak directions. Representation learning and feature lifts alter the geometry of the problem to make sure that informative variation appears earlier in the spectrum. Multimodal alignment and multitask learning can also be used to improve efficiency if tasks share underlying structure. These mechanisms may operate synergistically as well. Better representations reduce the number of modes that must be estimated, shifting the cross-over point where higher-capacity models outperform simpler ones. Additional data allow increasingly fine structure to be learned within the shared subspace. Conversely, if representations are poorly aligned with the underlying signal, increasing model complexity alone may not improve performance because additional capacity is applied to directions dominated by noise.

Viewed in this way, the success of modern machine learning methods depends less on model flexibility itself and perhaps mostly on how well representations concentrate signals into stable directions that can be estimated with finite data.

\subsection{Outlook}

Understanding how spectral structure interacts with model complexity may help guide method selection across biomedical domains. Rather than ask whether one model class is universally best, it may be more useful to match model complexity to the spectral structure of the data and the available sample size. Deep learning often succeeds when representations concentrate signals into stable directions that can be estimated reliably. The cross-over framework helps clarify when this advantage should emerge. Data efficiency can also improve through creative feature design.

\section{How to Use the Zeta Law in Practice}

The zeta law provides a practical framework for estimating how much data may be required for a given prediction problem, and whether collecting more data or improving the representation is likely to help most. Predictive performance depends on how disease-relevant signals accumulate across identifiable modes of variation in the representation used by the model.

A practical workflow has two components: empirical learning curves and spectral diagnostics.

First, estimate an empirical learning curve. Subsample the available data at several training set sizes $N$, train the model repeatedly, and evaluate performance using a relevant metric such as AUC, $R^2$, calibration error, or log likelihood. This produces an empirical curve $AUC(N)$ or $R^2(N)$ showing how performance improves as more data are added. Second, examine the spectrum of the representation produced by the encoder. Compute the covariance eigenspectrum of the features and, when applicable, the task-aligned signal spectrum.

For classification problems using linear discriminant analysis or related models, express the disease contrast in the covariance eigenbasis and examine how squared projections decay across modes. For cross-modal problems, compute singular values of the whitened cross-covariance operator

\begin{equation}
M
=
\Sigma_{xx}^{-1/2}
\Sigma_{xy}
\Sigma_{yy}^{-1/2},
\end{equation}

whose singular values quantify strength of shared structure between modalities. In many biological datasets, spectra approximately follow power laws over the range of modes that can be reliably estimated. If disease-aligned spectral energy decays approximately as

\begin{equation}
k^{-\beta},
\end{equation}

then cumulative detectable signal grows approximately as

\begin{equation}
\sum_{k=1}^{K(N)} k^{-\beta},
\end{equation}

where $K(N)$ is the number of modes that can be estimated reliably at sample size $N$.

When $\beta > 1$, most useful signal lies in early modes and performance tends to saturate relatively quickly.

When $\beta \approx 1$, improvements tend to follow approximately logarithmic growth.

When $\beta < 1$, signal is distributed across many weak modes and very large datasets may be required unless representations are improved. These diagnostics help determine which strategy may improve performance.

The power-law assumption does not need to hold globally. It is sufficient that the spectrum is approximately scale-free over the range of stable modes at available sample sizes.

These ideas may help guide study design. Rather than asking how much data is enough, investigators should really be asking how much data is enough for a given representation. Improved representations may greatly reduce sample requirements by concentrating biologically meaningful variation into fewer identifiable modes. Among counterintuitive predictions of the zeta law are the following hypotheses: adding weak modalities may improve performance; complex models may require less data when representations are effective; feature engineering influences sample complexity; and disease structure may often be low rank. In future work, we plan to test these predictions empirically by estimating spectral decay in learned representations and comparing predicted learning curves with observed improvements as sample size increases.

Large-scale biomedical consortia such as ENIGMA provide an ideal setting to test these models across modalities, disorders, and model classes.

\section{In Praise of Intellectual Creativity}

One implication of the zeta law is that creative representations may shift discovery curves as much as massive increases in sample size. In a way, that is why we are modeling data. Useful structure is often revealed by transforming the data. Weak features may become informative if combined, and redundant measures may stabilize important directions. Relational features such as correlations or gradients may outperform raw measures. Methods that appear to increase dimensionality, such as kernels, topological data analysis, or multimodal embeddings, may reduce effective dimensionality by concentrating signal into fewer stable modes. Auxiliary modalities, even when loosely informative, may act as coordinate systems revealing shared structure across datasets. The zeta law encourages creative feature lifts, geometric transformations, and multimodal alignments, all of them providing powerful strategies to improve discoverability by improving spectral structure.

\section*{Acknowledgments}

We thank the NIH for funding projects that motivated this work, and members of the ENIGMA consortium whose large-scale datasets and collaborative work enabled exploration of these ideas. We thank Dr Nikhil Dhinagar, Tamoghna Chattopadhyay, and Dr Yixue Feng for helpful discussions.


\begin{thebibliography}{99}

\bibitem{radford2021clip}
Radford, A., Kim, J. W., Hallacy, C., Ramesh, A., Goh, G., Agarwal, S., Sastry, G.,
Askell, A., Mishkin, P., Clark, J., Krueger, G., and Sutskever, I. (2021).
Learning transferable visual models from natural language supervision.
In \textit{Proceedings of the 38th International Conference on Machine Learning}, 8748--8763.

\bibitem{margulies2016situating}
Margulies, D. S., Ghosh, S. S., Goulas, A., Falkiewicz, M., Huntenburg, J. M.,
Langs, G., Bezgin, G., Eickhoff, S. B., Castellanos, F. X., Petrides, M.,
Jefferies, E., and Smallwood, J. (2016).
Situating the default-mode network along a principal gradient of macroscale cortical organization.
\textit{Proceedings of the National Academy of Sciences}, 113(44), 12574--12579.


\bibitem{dkw1956}
Dvoretzky, A., Kiefer, J., and Wolfowitz, J. (1956).
Asymptotic minimax character of the sample distribution function and of the classical multinomial estimator.
\textit{Annals of Mathematical Statistics}, 27(3), 642--669.

\bibitem{massart1990}
Massart, P. (1990).
The tight constant in the Dvoretzky--Kiefer--Wolfowitz inequality.
\textit{Annals of Probability}, 18(3), 1269--1283.

\bibitem{daviskahan1970}
Davis, C. and Kahan, W. M. (1970).
The rotation of eigenvectors by a perturbation. III.
\textit{SIAM Journal on Numerical Analysis}, 7(1), 1--46.

\bibitem{helmer2024cca}
Helmer, M., Warrington, S., Mohammadi-Nejad, A.-R., Ji, J. L., Howell, A., Rosand, B., and others (2024).
On the stability of canonical correlation analysis and partial least squares with application to brain–behavior associations.
\textit{Communications Biology}, 7, 217.
https://doi.org/10.1038/s42003-024-05869-4

\bibitem{saporta2024symile}
Saporta, A., Puli, A., Goldstein, M., and Ranganath, R. (2024).
Contrasting with Symile: Simple Model-Agnostic Representation Learning for Unlimited Modalities.
arXiv preprint arXiv:2411.01053.

\bibitem{dhinagar2026calmvlm}
Dhinagar, N. J., Jagad, C., Senthilkumar, P., Thomopoulos, S. I., Khan, M. H., 
Liew, S.-L., ENIGMA-Stroke Recovery Working Group, Banaj, N., Boric, M. R., 
Boyd, L. A., Brodtmann, A., Cassidy, J. M., Conforto, A. B., Cramer, S. C., 
Dula, A. N., Geranmayeh, F., Gregory, C. M., Hordacre, B., Jaywant, A., 
Kautz, S. A., Leech, K. A., Lotze, M., Mataró, M., Piras, F., Rosario, E. R., 
Sanossian, N., Schambra, H. M., Schweighofer, N., Seo, N. J., Soekadar, S. R., 
Thielman, G. T., Winstein, C., Wittenberg, G. F., Wong, K. A., and Thompson, P. M. (2026).
CALM-VLM: Calibration and selective prediction in vision–language models for reliable brain MRI classification.
\textit{bioRxiv}.
https://doi.org/10.64898/2026.04.10.717865


\bibitem{hettwer2022transdiagnostic}
Hettwer, M. D., Larivière, S., Park, B. Y., van den Heuvel, O. A., Schmaal, L.,
Andreassen, O. A., Ching, C. R. K., Hoogman, M., Buitelaar, J., van Rooij, D.,
Veltman, D. J., Stein, D. J., Franke, B., van Erp, T. G. M., ENIGMA ADHD Working Group,
ENIGMA Autism Working Group, ENIGMA Bipolar Disorder Working Group,
ENIGMA Major Depression Working Group, ENIGMA OCD Working Group,
ENIGMA Schizophrenia Working Group, Jahanshad, N., Thompson, P. M.,
Thomopoulos, S. I., Bethlehem, R. A. I., Bernhardt, B. C., Eickhoff, S. B., and Valk, S. L. (2022).
Coordinated cortical thickness alterations across six neurodevelopmental and psychiatric disorders.
\textit{Nature Communications}, 13, 6851.
https://doi.org/10.1038/s41467-022-34416-6


\bibitem{hettwer2026axes}
Hettwer, M. D., Saberi, A., Shafiei, G., Manoli, A., de Boer, A. A. A.,
van den Heuvel, O. A., Schmaal, L., Pozzi, E., Andreassen, O. A.,
Ching, C. R. K., Lawrence, K., Kim, G., Buitelaar, J., Turner, J. A.,
van Erp, T. G. M., Stein, D. J., Pine, D. S., Winkler, A. M.,
Bas-Hoogendam, J. M., Zugman, A., van der Wee, N. J. A., Groenewold, N. A.,
ENIGMA Autism Working Group, Calvo, R., Lázaro, L., Pariente, J.,
Behrmann, M., Dinstein, I., Haar, S., Ehrlich, S., King, J.,
Fair, D., Arango, C., Janssen, J., Parellada, M., Gonzalez Lois, N.,
Freitag, C. M., Daly, E., Murphy, D. G. M., Rubia, K., Ecker, C.,
Auzias, G., Deruelle, C., Hoekstra, L., Calderoni, S., Gori, I.,
Muratori, F., Retico, A., Tosetti, M., Bosco, P., Fedor, J.,
Luna, B., O'Hearn, K., Wallace, G. L., Jalbrzikowski, M.,
Rosa, P., Busatto, G., Martinho, M., Duran, F., Fitzgerald, J.,
Gallagher, L., McGrath, J., Anagnostou, E., Taylor, M., Kushki, A.,
Lerch, J., Durston, S., Oranje, B., Shook, D., van Horn, J. D.,
Jacokes, Z., McGartland Newman, B. F., Pelphrey, K. A.,
Bookheimer, S. Y., Dapretto, M., Webb, S. J., Jack, A.,
Nelson, C., Gaab, N., Bernier, R., Gupta, A., Bolte, S.,
Neufeld, J., Lundin Remnélius, K., Tammimies, K.,
ENIGMA Anxiety Working Group, Cano, M., Porta-Casteràs, D.,
Blair, J., Strawn, J., Steinhauser, J., Buckner, R., Soares, J.,
Brambilla, P., Straube, T., Price, R., Manfro, G., Mancini, M.,
Wittfeld, K., Mujica-Parodi, L., Klein, D., Fonzo, G.,
Assaf, M., Grotegerd, D., Cardoner, N., Blair, K.,
Schroeder, H., Khalsa, S., Nielsen, J., Maggioni, E.,
Critchley, H., van Nieuwenhuizen, H., Jin, F. J.,
Paulus, M., Diefenbach, G., Dannlowski, U.,
ENIGMA Bipolar Disorder Working Group, Pigoni, A.,
Vieta, E., Caseras, X., Temmingh, H., Goldstein, B.,
McDonald, C., Radua, J., Sim, K., Olie, E.,
Landén, M., Sacchet, M., Gotlib, I., Yatham, L. N.,
Yang, A. C., Janiri, D., Pomarol-Clotet, E.,
Scheffler, F., Ringin, E., Kennedy, K., Vilajosana, E.,
Clain, G., Klahn, L., Chi, I.-J.,
ENIGMA Major Depression Working Group, Portella, M. J.,
Schmidt, A., Hamilton, P., Okada, G., Walter, M.,
Besteher, B., Oudega, M., Dols, A., Whittle, S.,
Godlewska, B., Yang, T. T., Hill, D., Cullen, K.,
Vives-Gilabert, Y., Sempach, L., Koopowitz, S.,
Lake, M., Bauduin, S., Fuentes-Claramonte, P.,
Kamishikiryo, T., Li, M., Connolly, C., Grabe, H.,
Başgöze, Z., Völzke, H.,
ENIGMA OCD Working Group, Bruin, W., van Marle, H.,
van der Straten, A., Thomas, R., van Leeuwen, W.,
Benedetti, F., Calesella, F., Colombo, F.,
Kathmann, N., Beucke, J. C., Kaufmann, C.,
Brennan, B. P., Baker, J. T., Perriello, C.,
van Rooij, D., Brandeis, D., Cheng, Y., Xu, X.,
Xu, J., Jiang, L., Li, N., van Wingen, G.,
Gruner, P., Szeszko, P., Vriend, C.,
van der Werf, Y. D., Kasprzak, S., Hirano, Y.,
Nakagawa, A., Shimizu, E., Yoshida, T.,
Hoexter, M. Q., Batistuzzo, M. C., Sato, J. R.,
Koch, K., Rodriguez Manrique, D., Ruan, H.,
Kwon, J. S., Yun, J.-Y., Kim, M., Oh, H.,
Marsh, R., Fontaine, M., Mataix-Cols, D.,
Menchón, J. M., Alonso, P., Martínez-Zalacaín, I.,
Soriano-Mas, C., Bertolín, S., Morgado, P.,
Picó-Pérez, M., Fernandes, A., Silva Moreira, P.,
Sousa, N., Couto, B., Abe, Y., Sakai, Y.,
Feusner, J., O'Neill, J., Nurmi, E. L.,
Piacentini, J. C., Narayanaswamy, J. C.,
Venkatasubramanian, G., Balachander, S.,
Reddy, Y. C. J., Shivakumar, V., Simpson, H. B.,
Soreni, N., Minnuzi, L., Lochner, C., Ipser, J.,
Stern, E. R., Eng, G. K., Jaspers-Fayer, F.,
Stewart, E., Brem, S., Walitza, S., Wang, Z.,
Hu, H., Zhao, Q.,
ENIGMA Schizophrenia Working Group, Schall, U.,
Temmingh, H., Gruber, O., Calhoun, V.,
McKenna, P. J., Kircher, T., Nerland, S.,
Green, M., Sellgren, C. M., Ehrlich, S.,
Kochunov, P., Cobia, D., Crespo-Facorro, B.,
Banaj, N., Lebedeva, I., Borgwardt, S.,
Haukvik, U., Iasevoli, F., Van Rheenen, T.,
Kaiser, S., Picotin, R., Jensen, K., Stein, F.,
Quide, Y., Lee, M., Wang, L., Romero-Garcia, R.,
Vecchio, D., Tomyshev, A., Avram, M., Parker, N.,
Pontillo, G., Demir, A., Gonul, A. S.,
Hinc, A. C., Yazici, F., Sungur, I.,
Marquand, A., Bernhardt, B. C., Jahanshad, N.,
Thompson, P. M., Thomopoulos, S. I., Moore, T.,
Eickhoff, S. B., Kirschner, M., Satterthwaite, T. D.,
and Valk, S. L. (2026).
Brain alterations across mental disorders follow shared axes of cortical organization: a transdiagnostic ENIGMA study.
Manuscript in preparation.

\bibitem{ruan2026hodge}
Ruan, H., Chung, M. K., Bruin, W. B., Džinalija, N., Abe, Y., Alonso, P.,
Anticevic, A., Balachander, S., Batistuzzo, M. C., Benedetti, F., Bertolín, S.,
Brem, S., Cho, Y., Colombo, F., Couto, B., Eng, G. K., Ferreira, S., Feusner, J. D.,
Gruner, P., Hagen, K., Hansen, B., Hirano, Y., Hoexter, M. Q., Ipser, J.,
Jaspers-Fayer, F., Kim, M., Kwon, J. S., Lázaro, L., Li, C.-S. R., Lochner, C.,
Marsh, R., Martínez-Zalacaín, I., Menchón, J. M., Soriano-Mas, C., Morgado, P.,
Picó-Pérez, M., Fernandes, A., Sousa, N., Sakai, Y., O'Neill, J., Nurmi, E. L.,
Piacentini, J. C., Narayanaswamy, J. C., Venkatasubramanian, G., Reddy, Y. C. J.,
Shivakumar, V., Simpson, H. B., Soreni, N., Minnuzi, L., Stern, E. R.,
Stewart, E., Walitza, S., Wang, Z., Hu, H., Zhao, Q., Schall, U., Temmingh, H.,
Gruber, O., Calhoun, V., McKenna, P. J., Kircher, T., Nerland, S., Green, M.,
Sellgren, C. M., Ehrlich, S., Kochunov, P., Cobia, D., Crespo-Facorro, B.,
Banaj, N., Lebedeva, I., Borgwardt, S., Haukvik, U., Iasevoli, F.,
Van Rheenen, T., Kaiser, S., Picotin, R., Jensen, K., Stein, F., Quide, Y.,
Lee, M., Wang, L., Romero-Garcia, R., Vecchio, D., Tomyshev, A., Avram, M.,
Parker, N., Pontillo, G., Demir, A., Gonul, A. S., Hinc, A. C., Yazici, F.,
Sungur, I., Marquand, A., Bernhardt, B. C., Jahanshad, N., Thompson, P. M.,
Thomopoulos, S. I., and Eickhoff, S. B. (2026).
Disrupted higher-order topology in OCD brain networks revealed by Hodge Laplacian: an ENIGMA study.
\textit{bioRxiv}.
https://doi.org/10.64898/2026.03.04.709586

\end{thebibliography}
\end{document}